\documentclass[times,twocolumn,final,authoryear]{elsarticle}

\usepackage{framed,multirow}
\setcitestyle{square}

\usepackage{amsmath,lipsum}
\usepackage{amssymb}
\usepackage{latexsym}
\usepackage{textcomp}
\usepackage{siunitx}
\usepackage{etoolbox}

\usepackage{mathtools, cuted}

\usepackage{algorithmic}
\usepackage{stfloats}
\usepackage{xcolor}
\usepackage{lipsum}

\begin{document}

\thispagestyle{empty}

\begin{frontmatter}

\title{Graph Convolutional Networks: analysis, improvements and results}

\author[1]{Ihsan Ullah \corref{cor1}} 
\cortext[cor1]{Corresponding author: 
Tel.: +353-089-9897986;}
\ead{ihsan.ullah@nuigalway.ie}

\author[2]{Mario Manzo} 
\author[1]{Mitul Shah}
\author[1]{Michael Madden}

\address[2]{Information Technology Services, University of Naples "L'Orientale", Naples 80121, Italy}
\address[1]{Data Mining and Machine Learning Group, School of Computer Science, National University of Ireland Galway, Galway, Ireland}


\begin{abstract}
In the current era of neural networks and big data, higher dimensional data is processed for automation of different application areas. Graphs represent a complex data organization in which dependencies between more than one object or activity occur. Due to the high dimensionality, this data creates challenges for machine learning algorithms. Graph convolutional networks were introduced to utilize the convolutional models concepts that shows good results. In this context, we enhanced two of the existing Graph convolutional network models by proposing four enhancements. These changes includes: hyper parameters optimization, convex combination of activation functions, topological information enrichment through clustering coefficients measure, and structural redesign of the network through addition of dense layers. We present extensive results on four state-of-art benchmark datasets. The performance is notable not only in terms of lesser computational cost compared to competitors, but also achieved competitive results for three of the datasets and state-of-the-art for the fourth dataset.
\end{abstract}



\end{frontmatter}



\section{Introduction}
\label{introduction}
In the last few years, data is usually represented as points in a vector space. Today, structured data is omnipresent, able to include structural information between points and can be particularly important to better represent the models learned on them. For this purpose, the graphs are widely used to represent this type of information through nodes/vertices and edges, including local and spatial information derived from the data.
\par
Very often the interest concerns a prediction about the node properties in such graphs. For example, given a network that represents a human phenomenon, like a common exchange of messages in a social network, the goal is to predict the area of belonging i.e. users with common interests. Performing a forecasting process, especially in a semi-supervised environment, has been at the center of graph-based semi-supervised learning (\emph{SSL}) \cite{rozza2014novel}. In graph-based SSL, a small set of nodes is initially labeled. Starting from this information the remaining part of the graph structure is adopted, initially without the label, to label the nodes.
\par
Notationally, the structure described by the graph is normally incorporated as an explicit regularizer which applies a sliding constraint on the labels of the nodes to estimate. Recently, Graph Convolutional Networks (GCN)\cite{defferrard2016convolutional,kipf2016semi} have been proposed with a purpose to work on deep neural networks and graph-structured data. In this
paper, our attention is on the task of graph-based SSL using GCNs. GCN progressively estimate a transformation from graph to vector space, also called embedding, and an aggregation of neighborhood nodes, whereas a target loss
function for backpropagation errors is adopted. Indeed, node embedding result represents an estimation for label scores on the nodes. Moreover, it would be appropriate to obtain an association of confidence estimation to the label scores. This confidence scores can be adopted to understand the reliability of the estimated labels on a generic node. This improvement is introduced in the model called Confidence based Graph Convolutional Networks (ConfGCN) \cite{vashishth2019confidence}.
\par
In this context, the aim of our paper is fourfold:
\begin{enumerate}
    \item The models provide a set of parameters to be optimized. In this phase the goal is to find the optimal combination (activation function, loss function, hidden layers, number of nodes, etc) in order to obtain the best performance. It is assumed that this phase is very long and expensive.
    \item In past few years, many researchers have been worked designing novel activation functions in order to help deep neural networks to converge and obtain better performance. Standard neural networks employ logistic sigmoid activation functions which is affected by saturation problem and and consequently the effectiveness and efficiency of the classification is reduced. In this paper, we introduce an efficient approach to learn, during training, combinations of base activation functions (such as \textit{Relu6}); our goal is to check a search space for the activation functions through a convex combination of the base functions.
    \item The models adopt only the information relating to the degree of the single nodes (matrix $\tilde{D}$ in equation \ref{Hfun}) to process the graphs. In this regard, we introduce a measure that provides additional topological information called clustering coefficients.
    
    \item In deep learning the goal is to improve performance by focus attention on adding new layers, modifying the activation functions or changing the regularization methods. Furthermore, current structure, layers, can be redesigned to obtain optimal results compared to existing models. To this aim, we combined GCN and Dense layers. This new model provides a mixture of both GCN and Dense layers and compared to individual GCN, resulted in better performance.
    
\end{enumerate}
Additionally, we analyze the two baseline models, GCN and ConfCGN, in order to show the impact of proposed changes during training and testing phases. The paper is organized as follows: section \ref{sec:rel} gives an overview of related work. Section \ref{subsec:baselineModels} provides an overview of the two models GCN and ConfCGN. Section \ref{sec:proposedModels} explains the enhancement we proposed in this paper. Whereas, section \ref{sec:results} discusses the results achieved with proposed approach on GCN and confGCN. Finally, section \ref{sec:conclusion} gives some future directions and concludes our paper. 
\section{Related Work}
\label{sec:rel}
Recent literature provides some interesting insights about application of neural networks and data organized as graphs.
In \cite{kipf2016semi}  a variant of convolutional neural networks, called Graph   Convolutional Networks (GCNs), which operate directly on graphs is presented. Main motivation of convolutional architecture is related to localized first-order approximation of spectral graph convolutions. The model works by scaling linearly nodes connections and adopts hidden layer
representations which encode both structure and features of graphs. |||s on graphs, is presented. The proposed formulation does not alter the computational complexity of standard CNNs, despite being found to be processing graph structures. 

In \cite{marcheggiani2017encoding}, an enhanced version of \cite{kipf2016semi} is introduced. It is able to work with syntactic dependency graphs in form of sentence encoders and extracts words latent feature representations arranged in a sentence. Moreover, the authors showed that layers are complementary to LSTM layers.

In \cite{velickovic2018graph}, a neural network architecture for inductive and transductive problems, based on masked self-attentional layers, called graph attention networks (GATs), for graph-structured data is presented. In this model, nodes are able to contribute about neighboring features extraction and different weights to different nodes in a neighborhood are enabled, eliminating expensive matrix operations. In this way, several key challenges of spectral-based graph neural networks are addressed at the same time.

In \cite{vashishth2019confidence}, a modified version called Confidence-based Graph Convolutional Networks (ConfGCN) of \cite{kipf2016semi} is introduced. The improvement concerns a confidence estimation about label scores that has not been  explored in GCNs. ConfGCN adopts label scores estimation to identify the influence of a node on the neighborhood during aggregation, thus acquiring anisotropic abilities.

In \cite{liao2018graph} a graph partition neural networks (GPNN), an extension of graph neural networks (GNNs), useful to work with large graphs are described. GPNNs combine local information between nodes in small subgraphs and global information between the subgraphs. Graphs are partitioned in efficient way through several algorithms and, additionally, a novel variant for fast processing of large scale graphs is introduced. 

In \cite{yadav19a} a modified version of \cite{kipf2016semi} named Lov\'{a}sz Convolutional Networks (LCNs) is introduced. The model is able to capture global graph properties through Lov\'{a}sz orthonormal embeddings of the nodes.

In \cite{atwood2016diffusion}, a Diffusion-Convolutional Neural Networks (DCNNs) are described. Diffusion-convolution operation is useful to learn representations as an effective basis for node classification. The model includes different qualities such as latent representation for graphical data, invariance under isomorphism, polynomial-time prediction and learning. 
\par
In \cite{bruna2014spectral}, possible generalizations of Convolutional Neural Networks (CNNs) to signals is defined for more general domains. In particular, two models, one based upon a hierarchical clustering of the domain and another based on the spectrum of the graph Laplacian are described. The model is able to learn convolutional layers with a number of parameters independent of the input size, resulting in efficient deep architectures.
Further, a deep architecture with small learning complexity on general non-Euclidean domains is introduced in \cite{henaff2015deep}. The model is an extension of Spectral Networks which includes a graph estimation procedure.

Finally, in \cite{li2015gated} authors describe Gated graph sequence neural networks (GGNN), an extended versions of Graph Neural Networks (GNN) \cite{scarselli2008graph}, which uses modified gated recurrent units modern optimization techniques and extends output sequences. 
In the following section, we will explain the two baseline models i.e. GCN and ConfGCN. 
\section{Baseline Models}	
\label{subsec:baselineModels}
In this section, we briefly introduce Graph Convolutional Networks (GCNs) \cite{kipf2016semi} and its enhancement, Confidence-based Graph Convolutional Networks \cite{vashishth2019confidence}. The two models are compared and analyzed in detail in terms of their limitations and differences. Subsequently, starting from these points a set of improvements are proposed and demonstrated experimentally.
\subsection{Notations}
In this section we define the important elements for this research. Given $G=(V, E, X )$ an undirected graph, where $V = V_l \cup V_u$ the set containing labeled ($V_l$) and unlabeled ($V_u$)
nodes in the graph of dimension $n_l$ and $n_u$, $E$ is
the set of edges and $X \in \mathbb{R}^{(n_l+n_u) \times d}$ is the input node features. The label of a node $v$ is represented by a vector $Y_v \in \mathbb{R}^m$, belonging to $m$ classes. In this context, the goal is to predict the labels, $Y \in \mathbb{R}^{n_l \times m}$, of the unlabeled nodes of $G$. To consider the confidence, label distribution $\mu_v \in \mathbb{R}^m$ and a diagonal co-variance
matrix $\Sigma_{v} \in R_{m \times m}$ estimations are added, $\forall v \in V$. $\mu_{v,i}$ represents the score of label $i$ on node $v$, while $(\Sigma_{v})_{ii}$ represents the variance in the estimation of $\mu_{v,i}$. In other words, $(\Sigma_{v}^{-1})_{ii}$ is confidence in $\mu_{v,i}$.
\subsection{Graph Convolutional Networks}
Graph Convolutional Networks (GCNs) \cite{kipf2016semi} works on undirected graphs. Given a graph $G=(V, E,X)$, the node representation after a single layer of GCN can be defined as:
\begin{equation}
\label{Hfun}
    H=f((\tilde{D}^{-\frac{1}{2}}(A+I)\tilde{D}^{-\frac{1}{2}})XW)
\end{equation}
$W \in \mathbb{R}^{d \times d}$ includes the model parameters, $A$ represents nodes adjacency and $\tilde{D}_{ii}=\sum_{j}(A+I)_{ij}$. $f$ is any activation function such as $ReLU$, $f(x)=max(0,x)$. Equation \ref{Hfun} can be reformulated as:
\begin{equation}
    h_{v}=f\Biggl(\sum_{u \in N(v)}Wh_{u}+b\Biggl), \forall v \in V
\end{equation}
$b \in \mathbb{R}^{d}$ represents bias, $N(v)$ includes nodes neighborhood of $v$ in graph $G$ including $v$ and $h_{v}$ is representation of node $v$.
The goal is to acquire multi-hop dependencies between nodes,
different GCN layers can be superimposed over one
another. The representation of the node $v$ after $k$ layers can be written as
\begin{equation}
\label{hfun}
    h_{v}=f\Biggl(\sum_{u \in N(v)}(W^{k}h^{k}_{u}+b^{k})\Biggl), \forall v \in V
\end{equation}
where, $W^{k}$ and $b^{k}$ represent the weight and bias parameters of GCN layer.
\subsection{Confidence based Graph Convolutional Networks}
In \cite{vashishth2019confidence} Confidence-based Graph Convolutional Networks (ConfGCN) is described. The authors define the influence score of node $u$ considering its near node $v$ during GCN process as
follows:
\begin{equation}
\label{confval}
    r_{uv}=\frac{1}{d_{M}(u,v)}
\end{equation}
\begin{equation}
\label{mahalanobisdistance}
    {d_{M}(u,v)}=(\mu_{u}-\mu_{u})^T(\Sigma_{u}^{-1}+\Sigma_{v}^{-1})(\mu_{u}-\mu_{u})
\end{equation}
$d_{M}(u,v)$ represents Mahalanobis distance between two nodes \cite{orbach2012graph}. Specifically, considering nodes $u$ and $v$, with label distributions $\mu_{u}$ and $\mu_{v}$ and co-variance matrices $\Sigma_{u}$ and $\Sigma_{v}$, $r_{uv}$ gives more importance to spatially close nodes belonging to same class, otherwise reduces importance of nodes with low confidence scores. This results leads to inclusion of anisotropic capability during neighborhood exploration. For a node $v$, equation \ref{hfun} can be rewritten as:
\begin{equation}
\label{hfun2}
    h_{v}=f\Biggl(\sum_{u \in N(v)}r_{uv} \times (W^{k}h^{k}_{u}+b^{k})\Biggl), \forall v \in V.
\end{equation}
The final label prediction is obtained by equation \ref{ysoftmax} with $K$ number of layers.
\begin{equation}
\label{ysoftmax}
    \tilde{Y}_{v}=softmax(W^{K}h^{K}_{v}+b^{K}), \forall v \in V
\end{equation}
\subsection{GCN versus ConfGCN Models}
\label{Difference}
The differences between the two models are shown below:
\begin{enumerate}
    \item The major difference in both models is that GCN implements the nodes embedding, projection from graph space to  vector space, to describe the neighborhood while ConfGCN implements the confidence based prediction scheme where the neighboring nodes having higher confidence would be important parameters for the label of the unknown nodes.
    \item GCN model implements Chebychev polynomial method for the computational cost reduction while ConfGCN model uses the Loss Smoothening, regularization and optimization for better efficiency. Despite having more executional time per epoch ConfGCN model has better efficiency with the similar datasets.
    \item GCN doesn’t have constraints on the number of nodes that influences the representation of a given target node and each node is influenced by all the nodes in its k-hop neighborhood. Whereas, in ConfGCN the label confidences are used to ignore less confident nodes and nodes having higher confidence would be considered important. Furthermore, number of nodes influencing do not sway the prediction of the wrong labels.
    \item For more number of nodes in graphs such as Cora and CoraML datasets, ConfGCN has significantly better performances than Kipf GCN as the previous model implements the Node’s entropy of neighborhood calculation.
\end{enumerate}
The limitations of the two models are shown below:
\begin{enumerate}
    \item In GCN, memory requirement grows linearly in the size of the dataset. Whereas, ConfGCN requires higher memory requirement.
\item GCN is not applicable to directed graphs. It does not support edge features and is limited to undirected graphs (weighted or unweighted).
\item In GCN, locality for the nodes are assumed.
\item ConfGCN require more computational cost compared to the basic model. Cost increases as a confidence value, (equation \ref{confval}), for the exploration of the neighborhood node.
\item ConfGCN require more time for execution.
\item In ConfGCN, increasing layers reduces the accuracy. This behavior is connected to the increase of influencing nodes with increasing layers, which results in average information during aggregation.
\end{enumerate}



In the following section, we will explain the proposed enhancement we did for selecting an optimal deep model that may result in fewer executing time and enhanced performance. 
\section{Proposed Models: Enhanced GCN and ConfGCN}
\label{sec:proposedModels}
We proposed four major enhancement for both the models. The first enhancement is changing the hyper-parameters and training algorithm. The second and third are major enhancement i.e. adding more structural information to adjacency matrix and canonical optimization technique (also referred as convex). Finally, the fourth concerns a combination of two base models with introduction of additional dense layers. All these enhancement are applied on both the baseline models. Following section will explain how these models are designed and implemented. 
\subsection{Optimizing Hyper-parameters}
First we optimized the baseline models by fine-tuning the hyper-parameter that include activation function (AF), loss function (LF), and the number of nodes in each hidden layer. 
For AF we have explored it with $ReLu$, $ReLu6$, $Elu$, and $Selu$. In case of LF, we utilized simple cross entropy and cross entropy $softmax$ $V2$. Whereas, to explore the best number of nodes, we have taken nodes in each layer as $16$, $32$, $48$, $64$, $80$, $96$, $100$, $112$ and $200$. 
We explored to find the best combination of these parameters to provide optimal result in minimum amount of time. From now we will call the two enhanced versions Optimized Graph Convolutional Networks (OpGCN) and Optimized Confidence based Graph Convolutional Networks (OpConfGCN).
%
%
\subsection{Convex combination of activation functions}
A standard neural network $N_d$ can be composed of a set of hidden layers $d$ and a set functions $L_i$ that lead to a final mapping $\overline{L}$ related to a problem to address: $N_d=\overline{L} \circ L_d \circ \dots \circ L_1$. Specifically, each
hidden layer function $L_i$ is composed of two functions, $g_i$ and $\sigma_i$, which include parameters within the spaces $H_{gi}$ and $H_{\sigma i}$. A remapping of the layer input neurons in form of activation function can be seen as: $L_i = \sigma_i \circ g_i$. The learning process of $L_i$ consists in a procedure of optimization in the space $H_i = H_{\sigma i} \times H_{gi}$. Commonly, $\sigma_i$ does not provide for a learning phase and $H_{\sigma i}$ is a singleton. Then, $H_i = \{\sigma_i\} \times H_{gi}$. 
If we consider a fully-connected layer from $\mathbb{R}^{n_i}$ to
$\mathbb{R}^{m_i}$ which adopts Relu activation function, $H_{gi}$ represents the set of all
affine transformations from $\mathbb{R}^{n_i}$ to
$\mathbb{R}^{m_i}$, then $H_i= {ReLu} \times Lin(\mathbb{R}^{n_i},\mathbb{R}^{m_i}) \times K(\mathbb{R}^{m_i})$, where $Lin(A,B)$ and $K(B)$ are respectively the sets of linear maps between $A$ and $B$, and the set of translations of $B$.
\par
In this paper, we adopt a technique to define learnable activation functions \cite{manessi2018learning} that can be adopted in all hidden layers of a GCN architecture. 
%
The approach consists of a hypothesis space $H_{\sigma i}$ and is based on the following idea: 
\begin{itemize}
    \item select a set of activation functions $F= \{f_1,\dots,f_N\}$, in which elements can be adopted as base elements;
    \item fix the activation function $\sigma_i$ to combine in linear way as elements belonging to $F$ set;
    \item look for an optimal hypothesis space;
    \item look for GCN optimization respect to $H_i = H_{\sigma i} \times H_{gi}$.
\end{itemize}
%

\begin{equation}
conv(A) := \{\Sigma_i c_i a_i| \Sigma_i c_i=1, c_i  \ge  0, a_i \in A\};    
\end{equation}

$conv(A)$ is not vector subspace of $V$ and is a generic convex subset in $V$ reducing to a $(|A|-1)$-dimensional simplex when
the elements of $A$ are linearly independent. If we consider $F:=\{f_0, f_1,\dots, f_N\}$ the set of activation functions $f_i$, the vector space \textbf{\textit{F}} is defined from F considering all linear combinations $\sum_i c_i f_i$ with $c_i  \ge  0, \Sigma_i c_i=1$. Note
that, despite $F$ is a spanning set of $\textbf{\textit{F}}$, it is not generally a basis; indeed $|F| \ge dim$ $\textbf{\textit{F}}$. Based on previous definitions, we can now define the  technique to build
learnable activation functions as follows:
\begin{itemize}
    \item fix a finite set $F = \{f_1,\dots, f_N \}$, where each $f_i$ is a learneable activation function;
    \item create an additional activation function $\overline{f}$ as a linear combination of all the $f_i \in F$;
\item select as hypothesis space $H_{\overline{f}}$ the $conv(F)$ set;

\end{itemize}

The results are obtained for the following combination for $F$:

\begin{equation}
    F:=\{Relu6,Relu6\}
\end{equation}

where

\begin{equation}
    Relu6=min(max(0,x),6)
\end{equation}

In this work we have implemented two methods:
\begin{enumerate}
    \item Taking two input layers of the model, use the different activation for them and then applying any mathematical operations on the inputs, i.e. Summation, Subtraction, Maximum, minimum and Average values of both Input layer’s output.
    \item  Looking at those results we got to know that summation operations are having the best results so we applied the canonical form on the outputs. In this case the convex combination becomes $conv(A) := c_1Relu6+c_2Relu6$. The Structure of Base-Line model with optimized results is shown in Table \ref{tab-aa}:
\end{enumerate}

\begin{table}[!ht]
\caption{Baseline Model structure for enhancing with convex approach}
\label{tab-aa}
\centering
\scriptsize
\begin{tabular}{|c|c|c|c|c|}
\hline
Input Size & L1-Nodes & L1-ActivationFun & OutputNodes & loss function \\
\hline
1433 & 16 & Relu & 3 & Cross Entropy \\
\hline
\end{tabular}
\end{table}
Whereas its enhanced model structure is given in Table \ref{tab-ab}:

\begin{table}[!ht]
\caption{Enhanced Model structure for convex approach}
\label{tab-ab}
\centering
\tiny
\begin{tabular}{|c|c|c|c|c|c|c|c|}
\hline
 In-Size & L1-Nodes & L1a-AF & L1b-AF & Out-Nodes & LossFun & $c_1$ & $c_2$\\
\hline
1433 & 16 & Relu6 & Relu6 & 3 & CrossEntropy& 0.8 & 0.2\\
\hline
\end{tabular}
\end{table}

From now we will call the two enhanced versions Convex Graph Convolutional Networks (ConvGCN) and Convex Confidence based Graph Convolutional Networks (ConvConfGCN).
\subsection{Clustering coefficients}
\label{subsec:clustcoeficient}
In equation \ref{Hfun} the adjacency matrix $A$, which describes the topology of the network, is very significant part of both models. Furthermore, the identity matrix $I$ is added to $A$ in order to remove zero values on the main diagonal. Our idea is to add more information about nodes by introducing a particular property called Clustering Coefficients. In graph theory, the clustering coefficient describes the degree of aggregation of nodes in a graph. The measure is based on triplets of nodes. A triplet is defined as three connected nodes. A triangle can include three closed triplets, each one centered on one of the nodes. Two possible versions can be defined: the Global Clustering Coefficients (GCCs) and the local Clustering Coefficients (CCs) \cite{opsahl2013triadic}. We adopt the second defined as:

\begin{equation}
    CC_i=\frac{\delta_i}{k_i(k_i-1)}
\end{equation}

$k_i$ is the degree of node $i$ and $\delta_i$ is the number of edges between the $k_i$ neighbors of node $i$. The measure is in the range $\{0,\ldots,1\}$, $0$ if none of the neighbors of a node is connected and $1$ if all of the neighbors are connected. Topological information is provided through CCs, which is connected to other structural properties \cite{strang2018generalized}, such as transitivity, density, characteristic path length, and efficiency, useful for representation in the vector space. In this work we are suggesting that there is another possibility of the matrix $I$ which is to replace the main diagonal of the matrix $I$ with CCs values. For a graph of $n \times n$ nodes the identity matrix becomes: 
\begin{equation}
   I_n = \begin{bmatrix}
CC_1 & 0 & 0 & \cdots & 0 \\
0 & CC_2 & 0 & \cdots & 0 \\
0 & 0 & CC_3 & \cdots & 0 \\
\vdots & \vdots & \vdots & \ddots & \vdots \\
0 & 0 & 0 & \cdots & CC_n \end{bmatrix}
\end{equation}
From now we will call the two enhanced versions Clustering Coefficients Graph Convolutional Networks (CCGCN) and Clustering Coefficients Confidence based Graph Convolutional Networks (CCConfGCN).
The Structure of Base-Line model with optimized results is similar to Table \ref{tab-aa}. 
Matrix was added to the Adjacency matrix while pre-processing of the input and the combined matrix was considered as input to the neural network. The new matrix $I_n$, having the same size as Identity matrix, is added to the adjacency matrix instead of plain identity matrix.  
%
%
\subsection{GCN and Dense Layer combination}
\label{sec:mixGCNDense}
Deep learning models have shown that beside creating a new layer, activation function, regularization method etc., if one can redesign existing layers etc. in a proper way. It can result in optimal performance as compare to the previous models. We adopted the same GCN and Dense layers and created a model that gave the optimal results. A dense layer is commonly known as fully connected layer and it is represented as: 
\begin{equation}
\label{eq:fullconOutput}
\resizebox{0.3\textwidth}{!}{$
y_{u}^{l_{n}} = f_{l_{n}} \left ( \sum_{i=1}^{I} \left ( \left ( w_{(i,v)}^{l_{n}} ~.~  y_{(i)}^{l_{n-1} } \right ) + b_{(1,v)}^{l_{n} } \right ) \right)
$}
\end{equation}
Here, $y_{u}^{l_{n}}$ represents the neuron at layer $n$, $w_{i,v}^{l_{n}}$ represents the weight $(i,v)$ for that neuron multiplied with input neuron $y_{i}^{l_{n-1}}$, and $b_{v}^{l_{n}}$  represents that bias that is added to the weighted sum. The resultant weighted sum value is passed through an activation function $f_{l_{n}}$.
Table \ref{tabmixgcndense} shows the structure of the model. We used this model on all four datasets. After extensive experiments, the best results are shown in Table \ref{tab2}. This combination provides a mixture of both GCN and Dense layers and result in better performance compared to individual GCN or Dense layer.

The training phase adopts the same Adam optimizer (similar to all other models). In each layer, we used \textit{Relu6} activation function. From now we will call the two enhanced versions Dense Graph Convolutional Networks (DGCN) and Dense Confidence based Graph Convolutional Networks (DConfGCN).

\begin{table}[!ht]
\caption{Model having both GCN and Dense Layer}
\label{tabmixgcndense}
\centering
\scriptsize
\centering
\begin{tabular}{|c|c|c|c|c|c|c|c|}
\hline
Layer & In-Nodes & Out-Nodes & AF & DO  \\
\hline
Input & 1433 & - & - & - \\
\hline
GCN & 1433 & 32 & Relu6 & 0.5 \\
\hline
Dense-1 & 32 & 16 & Relu6 & 0.5 \\
\hline
Dense-2 & 16 & 32 & Relu6 & 0.5\\
\hline
GCN & 32 & 48 & Relu6 & 0.5 \\
\hline
GCN & 48 & 7 & - & 0.5 \\
\hline
Output & 7 & 7 & Softmax & - \\\hline
\end{tabular}
\end{table}

In Table \ref{tabmixgcndense}, 'In-Nodes' represents the input nodes to a layer, 'Out-Nodes' represent the output nodes of a layer, 'AF' represents the activation function, whereas drop out rate is represented by 'DO'. 

\section{Results}
\label{sec:results}
This section describes the results obtained on public datasets with the proposed improvements. In addition, the achieved results will be compared to the state-of-the-art models in literature. 
\subsection{Datasets}
\label{datasets}
For performance evaluation we adopt several semi-supervised classification datasets that are commonly used by other researchers. The set of dataset comprise of Cora, Citeseer, Pubmed \cite{sen2008collective}, and Cora-ML \cite{bojchevski2018deep}. The setup is the same as being followed in \cite{vashishth2019confidence}.
Our aim concerns to classify documents into one of the predefined classes. Datasets represent citation networks, in which each document is encoded using bag-of-words features with undirected edges between nodes. The dataset statistics is summarized in table \ref{tab1}. Label mismatch concerns the fraction of edges between nodes with different labels in the training data. The datasets have substantially low label mismatch rate except Cora-ML.
\begin{table}[!ht]
\caption{Dataset statistics.}
\label{tab1}
\centering
\scriptsize
\begin{tabular}{|c|c|c|c|c|c|c|}
\hline
Dataset & Nodes & Edges & Classes & 
Features & Labels Mismatch & $\frac{V_l}{V}$\\
\hline
Cora & 2708  & 5429 & 7 & 1433 & 0.002  & 0.052\\
\hline
Cora-ML & 2995 & 8416 &  7 & 2879 & 0.018 & 0.166\\
\hline
Citeseer & 3327 & 4372 & 6 & 3703 & 0.003 & 0.036\\
\hline
Pubmed & 19717 & 44338 & 3 & 500 & 0.0 & 0.003\\
\hline
\end{tabular}
\end{table}

\subsection{Competitors}
\label{Competitors}

Our method is compared with approaches of different nature. Competitors can be divided into four groups. First group includes approaches based on extensions of the GCN model. G-GCN \cite{marcheggiani2017encoding} provides an extension adopting edge-wise gating to remove noisy edges during aggregation. GAT \cite{velickovic2018graph} provides a method based on attention which gives different weights to different nodes by allowing nodes to attend to their neighborhood. Dual-GCN \cite{monti2018dual} allows to learn both vertex and edge features and generalizes the GAT model \cite{velickovic2018graph}. LGCN \cite{gao2018large} works based on a learnable graph convolutional layer (LGCL). LGCL automatically selects a fixed number of neighboring nodes for each feature based on value ranking in order to transform graph data into grid-like structures in 1-D format. Fast-GCN \cite{liang2015fastgcn} is an accelerated and optimized tool for constructing gene co-expression networks that can fully harness the parallel nature of GPU (Graphic Processing Unit) architectures. Second group includes approaches based on extensions of the GNN model \cite{scarselli2008graph}. GGNN \cite{li2015gated} generalizes RNN framework for graph-structured data application. GPNN \cite{liao2018graph} adopts partition approach to spread the information after the subdivision of large graphs into subgraphs. Third group includes approaches based on embedding. SemiEmb \cite{weston2012deep} is a framework which provides semi-supervised regularization to improve training. DeepWalk \cite{perozzi2014deepwalk} adopts random walks to learns node features. Planetoid \cite{yang2016revisiting} adopts a transductive and inductive approach for class label prediction using neighborhood information. Fourth group includes baseline approaches. LP \cite{zhu2003semi} is a label propagation algorithm which spreads labels information to neighborhood following the proximity. ManiReg \cite{belkin2006manifold} provides geometric regularization on data. Feat \cite{yang2016revisiting} works based on node features ignoring the structure information.



\subsection{Comparison}
\label{setup}
 We have summarized our results by showing the best results of all the enhancements for all the datasets. Table \ref{tab2} shows the accuracy of all the models mentioned in Section \ref{sec:proposedModels}. We have been successful in getting state-of-the-art result on one dataset as well as very close to the state-of-the-art work done till now on the other three datasets as highlighted in green in Table \ref{tab2}. 
On Cora\_ML dataset we achieved the current best accuracy of $86.9\pm0.4$ using DConfGCN model. This is the current state-of-the-art based on our knowledge as the most recent papers i.e. Dual-GCN, LGCN, and Fast-GCN did not reported their results on Cora\_ML dataset. 
In case of ‘Citeseer’ dataset, the best result which we achieved is 73.26\%, this is more than Dual GCN and 0.3\% less from LGCN. This makes our accuracy with ConvConfGCN the second best till date. However, just to highlight that LGCN \cite{gao2018large} report only the best result whereas our result are based on 100 run which are more stronger compare to reporting one highest performance.
We have got the $3^{rd}$ best accuracy for ‘Pubmed’ dataset  i.e. $79.8 \pm 0.4$. Finally, on ‘Cora’ dataset, we achieved $82.1 \pm 1.2$ accuracy with DGCN that is better than baseline GCN and ConfGCn by slight margin, but at $4^{th}$  position overall in the list. 

\par
One of the reason for not having the best result for Citeseer, Cora, and Pubmed could be that the best reported results in the papers \cite{gao2018large, monti2018dual, liang2015fastgcn} are not having the mean performance over multiple runs. Another reason is that, our model can not be directly compared with model like LGCN as it uses regular convolutional kernel in their model. Rather designing new kernels to work on graph data, in LGCN the authors organized the graph data in a way that normal convolutional kernel can operate over it and learn feature from them. 
These enhancement and results are reported to provide baseline for future works to be done in the field of SSL for the Graphs. 

\begin{table}[!ht]
\caption{Performance comparison of different methods on described datasets.}
\label{tab2}
\centering
\tiny
\begin{tabular}{|c|c|c|c|c|}
\hline
\textbf{Method} & \textbf{Citeseer} & \textbf{Cora} & \textbf{Pubmed} & \textbf{Cora ML} \\
\hline
LP \cite{zhu2003semi} & 45.3 & 68.0 & 63.0 & -\\
\hline
ManiReg \cite{belkin2006manifold} & 60.1 & 59.5 & 70.7 & -\\
\hline
SemiEmb \cite{weston2012deep} & 59.6 & 59.0 & 71.1 & -\\
\hline
Feat \cite{yang2016revisiting} & 57.2 & 57.4 & 69.8 & -\\
\hline
DeepWalk \cite{perozzi2014deepwalk} & 43.2 & 67.2 & 65.3 & -\\
\hline
GGNN \cite{li2015gated} & 68.1 & 77.9 & 77.2 & -\\
\hline
Planetoid \cite{yang2016revisiting} & 64.9 & 75.7 & 75.7 & -\\
\hline
G-GCN \cite{marcheggiani2017encoding} & 69.6 $\pm$ 0.5 & 81.2 $\pm$ 0.4 & 77.0 $\pm$ 0.3 & 86.0 $\pm$ 0.2\\
\hline
GPNN \cite{liao2018graph} & 68.1 $\pm$ 1.8 & 79.0 $\pm$ 1.7 & 73.6 $\pm$ 0.5 & 69.4 $\pm$ 2.3\\
\hline
GAT \cite{velickovic2018graph} & 72.5 $\pm$ 0.7 & 83.0 $\pm$ 0.7 & 79.0 $\pm$ 0.3 & 83.0 $\pm$ 0.8\\
\hline
GCN \cite{kipf2016semi} & 69.4 $\pm$ 0.4 & 80.9 $\pm$ 0.4 & 76.8 $\pm$ 0.2 & 85.7 $\pm$ 0.3 \\
\hline
\textbf{OpGCN} & 70.1$\pm$ 0.7 & 80.3$\pm$ 0.4 & 79.1$\pm$ 0.3 & 85.3$\pm$ 0.4\\
\hline
\textbf{ConvGCN} & 70.1 $\pm$ 0.3 & 80.1$\pm$ 0.2 & 79.0$\pm$ 0.2 & 84.3$\pm$ 0.3 \\
\hline
\textbf{CCGCN} & 53.1 $\pm$ 0.6 & 55.3 $\pm$ 2.4 & 71.1 $\pm$ 0.7 & 63.3 $\pm$ 0.4  \\
\hline
\textbf{DGCN} & 70.9 $\pm 0.7$ & \textcolor{green}{\textbf{$82.1 \pm 1.2$}} & $79.10 \pm 0.4$ & $86.3 \pm 0.3$  \\
\hline
ConfGCN \cite{vashishth2019confidence} & 72.7 $\pm$ 0.8 & 82.0
$\pm$ 0.3 & 79.5 $\pm$ 0.5 & 86.5 $\pm$ 0.3\\
\hline
\textbf{OpConfGCN} & 70.1 $\pm$ 1.4
 & 80.9 $\pm$ 0.8
 & \textcolor{green}{\textbf{79.8 $\pm$ 0.4}}
 & 84.6 $\pm$ 0.5\\
 \hline
 \textbf{ConvConfGCN} & \textcolor{green}{\textbf{73.1$\pm$ 0.2}} & 82.1$\pm$ 0.6 & \textcolor{green}{\textbf{79.8$\pm$ 0.4}} & 86.4$\pm$ 0.3\\
\hline
\textbf{CCConfGCN} & 70.8 $\pm$ 0.3
& 82.1 $\pm$ 0.6
 & 78.2 $\pm$ 0.4 & 83.4 $\pm$ 0.5\\
\hline
\textbf{DConfGCN} & 58.03 $\pm$ 0.9 & 81.0 $\pm$ 1.4 & 78.8 $\pm$ 0.6 & \textcolor{green}{\textbf{86.9 $\pm$ 0.4}} \\
\hline
Dual-GCN \cite{monti2018dual} & 72.6 & 83.5 & 80.0 & - \\
\hline
LGCN \cite{gao2018large} & 73.4 & 83.3 & 79.7 & -\\
\hline
Fast-GCN \cite{liang2015fastgcn} & - & 86 & 88 & - \\
\hline
\end{tabular}
\end{table}

In table \ref{time} execution time for PubMed dataset is shown. As the size of the features in each dataset varies, that is why the time (in seconds) per epoch varies for each dataset. GCN and its enhancements are faster compared to confGCN and its enhancements. While optimizing based on hyper-parameters, we found that the major reduction in computational cost was due to usage of the cross-entropy softmax V2 function rather than simple cross-entropy. Therefore, in all our later experiments we used this loss function. ConfGCN based models took more time compare to GCN based models. The optimal models interms of execution time is OPGCN. 
\begin{table}[!ht]
\caption{Execution time on Pubmed dataset}
\label{time}
\centering
\begin{tabular}{|c|c|}
\hline
Method & Time (sec) \\
\hline
GCN \cite{kipf2016semi} & 0.8 \\
\hline
\textbf{OpGCN} & 0.415\\
\hline
ConvGCN  & 0.585\\
\hline
CCGCN  & 0.417 \\
\hline
DGCN  & 0.662 \\
\hline
ConfGCN \cite{vashishth2019confidence} & 1.344\\
\hline
OpConfGCN & 1.93\\
 \hline
ConvConfGCN & 1.96\\
\hline
CCConfGCN & 1.93\\
\hline
DConfGCN & 1.99\\
\hline
\end{tabular}
\end{table}

\section{Conclusions}
\label{sec:conclusion}
We present enhanced models of GCN and ConfGCN for the Graph Convolution with Semi-supervised learning. The main focus among all the enhancements was on four changes: parametric configuration, adding more structural information to adjacency matrix for graph representation, convex optimization related to activation functions and combination of base models and dense layers. As this work is related to the Graph convolutions and with these enhanced models we have been able to show that the addition of the layers can be helpful for the increment of the accuracy, so this process has opened a path where ‘addition of layer means accuracy reduction’ limitation of SSL has been removed. Also currently all the Graph Convolutional Layers are using 1D convolutions to operate the model, there can be 2D or 3D dimensional weighing schemes can be implemented on the concurrent models. The GCN was a new approach for SSL and in that the layer-wise propagation rule was implemented while ConfGCN is a model which estimates label scores with labels’ confidence.
We have prepared six different models with different configurations and we have validated our models with four benchmark datasets. In majority of the enhancements, we have been successful in increasing the accuracy as well as the execution time for all the best possible configurations in all four data-sets.



\section*{Acknowledgments}
The first two authors acknowledge the guidance and supervision of their late Prof. Alfredo Petrosino. May he rest in peace.  

\bibliographystyle{plainnat}
\bibliography{refs}

\end{document}